\documentclass[journal,twoside]{IEEEtran}

\usepackage{amsmath,amssymb,bm,cite,graphicx,hyperref,multirow,url}
\usepackage[caption=false,font=footnotesize]{subfig}

\newcommand{\R}{\mathbb{R}}
\newcommand{\x}{\mathbf{x}}
\newcommand{\z}{\mathbf{z}}

\begin{document}

\title{Frank-Wolfe Network: An Interpretable Deep Structure for Non-Sparse Coding}

\author{Dong~Liu,~\IEEEmembership{Senior Member,~IEEE,}
  Ke~Sun,
  Zhangyang~Wang,~\IEEEmembership{Member,~IEEE,}\\
  Runsheng~Liu,
  and~Zheng-Jun~Zha,~\IEEEmembership{Member,~IEEE}
  \thanks{Date of current version \today. This work was supported by the National Key Research and Development Plan under Grant 2017YFB1002401, and by the Natural Science Foundation of China under Grant 61772483.

  D. Liu, K. Sun, and Z.-J. Zha are with the CAS Key Laboratory of Technology in Geo-Spatial Information Processing and Application System, University of Science and Technology of China, Hefei, China (e-mail: dongeliu@ustc.edu.cn; sunk@mail.ustc.edu.cn; zhazj@ustc.edu.cn).

  Z. Wang is with Department of Computer Science and Engineering, Texas A\&M University, College Station, TX, USA (e-mail: atlaswang@tamu.edu).

  R. Liu was with the University of Science and Technology of China, Hefei, China (e-mail: lrs1892@outlook.com).

  Copyright \copyright~2019 IEEE. Personal use of this material is permitted. However, permission to use this material for any other purposes must be obtained from the IEEE by sending an email to pubs-permissions@ieee.org.}
  }

\markboth{IEEE Transactions on Circuits and Systems for Video Technology}%
{Liu \MakeLowercase{\textit{et al.}}: Frank-Wolfe Network: An Interpretable Deep Structure for Non-Sparse Coding}

\IEEEtitleabstractindextext{%
\begin{abstract}
The problem of $L_p$-norm constrained coding is to convert signal into code that lies inside an $L_p$-ball and most faithfully reconstructs the signal. Previous works under the name of sparse coding considered the cases of $L_0$ and $L_1$ norms. The cases with $p>1$ values, i.e. non-sparse coding studied in this paper, remain a difficulty. We propose an interpretable deep structure namely Frank-Wolfe Network (F-W Net), whose architecture is inspired by unrolling and truncating the Frank-Wolfe algorithm for solving an $L_p$-norm constrained problem with $p\geq 1$. We show that the Frank-Wolfe solver for the $L_p$-norm constraint leads to a novel closed-form nonlinear unit, which is parameterized by $p$ and termed $pool_p$. The $pool_p$ unit links the conventional pooling, activation, and normalization operations, making F-W Net distinct from existing deep networks either heuristically designed or converted from projected gradient descent algorithms. We further show that the hyper-parameter $p$ can be made learnable instead of pre-chosen in F-W Net, which gracefully solves the non-sparse coding problem even with unknown $p$. We evaluate the performance of F-W Net on an extensive range of simulations as well as the task of handwritten digit recognition, where F-W Net exhibits strong learning capability. We then propose a convolutional version of F-W Net, and apply the convolutional F-W Net into image denoising and super-resolution tasks, where F-W Net all demonstrates impressive effectiveness, flexibility, and robustness.
\end{abstract}

}

\maketitle

\IEEEdisplaynontitleabstractindextext

\IEEEpeerreviewmaketitle

\section{Introduction}
\label{sec_introduction}
\subsection{$L_p$-Norm Constrained Coding}
Assuming a set of $n$-dimensional vectors $\{\x_i\in\R^n|i=1,2,\dots,N\}$, we aim to encode each vector $\x_i$ into a $m$-dimensional code $\z_i\in\R^m$, such that the code reconstructs the vector faithfully. When $m>n$ is the situation of interest, the most faithful coding is not unique. We hence consider the code to bear some structure, or in the Bayesian language, we want to impose some prior on the code.
One possible structure/prior is reflected by the $L_p$-norm, i.e. by solving the following problem:
\begin{eqnarray}\label{pncc-known-p}
  \{\{\z_i^*\},D^*\}=&\arg\min{\sum_{i=1}^{N}{\|\x_i-D\z_i\|^2_2}}\nonumber \\
  &\mbox{subject to }\forall i, \|\z_i\|_p\leq c
\end{eqnarray}
where $D\in\R^{n\times m}$ is a linear decoding matrix, and both $p$ and $c$ are constants. In real-world applications, we have difficulty in pre-determining the $p$ value, and the flexibility to learn the prior from the data becomes more important, which can be formulated as the following problem:
\begin{eqnarray}\label{pncc-unknown-p}
  \{\{\z_i^*\},D^*,p^*\}=&\arg\min{\sum_{i=1}^{N}{\|\x_i-D\z_i\|^2_2}}\nonumber \\
  &\mbox{subject to }p\in\mathcal{P},\forall i, \|\z_i\|_p\leq c
\end{eqnarray}
where $\mathcal{P}\subset\R$ is a domain of $p$ values of interest. For example, $\mathcal{P}=\{p|p\geq 1\}$ defines a domain that ensures the constraint to be convex with regard to $\z_i$. In this paper, (\ref{pncc-known-p}) and (\ref{pncc-unknown-p}) are termed $L_p$-norm constrained coding, and for distinguishing purpose we refer to (\ref{pncc-known-p}) and (\ref{pncc-unknown-p}) as known $p$ and unknown $p$, respectively.

\subsection{Motivation}
For known $p$, if the decoding matrix is given a priori as $D^*=D_0$, then it is sufficient to encode each $\x_i$ individually, by means of solving:
\begin{eqnarray}\label{lp-cons}
  \z_i^*=&\arg\min{\|\x_i-D_0\z_i\|^2_2}\nonumber \\
  &\mbox{subject to }\|\z_i\|_p\leq c
\end{eqnarray}
or its equivalent, unconstrained form (given a properly chosen Lagrange multiplier $\lambda$):
\begin{equation}\label{lp-reg}
\z_i^*=\arg\min{\|\x_i-D_0\z_i\|^2_2+\lambda\|\z_i\|_p}
\end{equation}
Eq. (\ref{lp-reg}) is well known as an $L_p$-norm regularized least squares problem, which arises in many disciplines of statistics, signal processing, and machine learning. Several special $p$ values, such as 0, 1, 2, and $\infty$, have been well studied. For example, when $p=0$, $\|\z_i\|_0$ measures the number of non-zero entries in $\z_i$, thus minimizing the term induces the code that is as \emph{sparse} as possible. While sparsity is indeed a goal in several situations, the $L_0$-norm minimization is NP-hard \cite{natarajan1995sparse}. Researchers then proposed to adopt $L_1$-norm, which gives rise to a convex and easier problem, to approach $L_0$-norm. It was shown that, under some mild conditions, the resulting code of using $L_1$-norm coincides with that of using $L_0$-norm \cite{donoho2006most}. $L_1$-norm regularization was previously known in lasso regression \cite{tibshirani1996regression} and basis pursuit \cite{chen2001atomic}. Due to its enforced sparsity, it leads to the great success of compressive sensing \cite{donoho2006compressed}. The cases of $p>1$ lead to non-sparse codes that were also investigated. $L_2$-norm regularization was extensively adopted under the name of weight decay in machine learning \cite{krogh1992simple}. $L_{\infty}$-norm was claimed to help spread information evenly among the entries of the resultant code, which is known as democratic \cite{studer2014democratic} or spread representations \cite{fuchs2011spread}, and benefits vector quantization \cite{lyubarskii2010uncertainty}, hashing  \cite{wang2017transformed}, and other applications.

Besides the above-mentioned special $p$ values, other general $p$ values were much less studied due to mathematical difficulty. Nonetheless, it was observed that general $p$ indeed could help in specific application domains, where using the special $p$ values might be overly simplified. For sparse coding, $L_p$-norms with $p\in[0,1]$ all enforce sparsity to some extent, yet with different effects. In compressive sensing, it was known that using $L_p$-norm with $0<p<1$ needs fewer measurements to reconstruct sparse signal than using $L_1$-norm; regarding computational complexity, solving $L_p$-norm regularization is more difficult than solving $L_1$-norm but still easier than solving $L_0$-norm \cite{chartrand2007exact}. Further studies found that the choice of $p$ crucially affected the quality of results and noise robustness \cite{chartrand2008restricted}. Xu \emph{et al.} endorsed the adoption of $L_{1/2}$-norm regularization and proposed an iterative half thresholding algorithm \cite{xu2012l12}. In image deconvolution, Krishnan and Fergus investigated $L_{1/2}$ and $L_{2/3}$ norms and claimed their advantages over $L_1$-norm \cite{krishnan2009fast}. For non-sparse coding, $L_p$-norms with $p>1$ and different $p$'s have distinct impact on the solution. For example, Kloft \emph{et al.} argued for $L_p$-norm with $p>1$ in the task of multiple kernel learning \cite{kloft2011lp}.

Now let us return to the problem (\ref{pncc-known-p}) where the decoding matrix $D$ is \emph{unknown}. As discussed above, $L_0$-norm induces sparse representations, thus the special case of $L_p$-norm constrained coding with $p=0$ (similarly $p=1$) is to pursue a sparse coding of the given data. While sparse coding has great interpretability and possible relation to visual cognition \cite{olshausen1997sparse,lee2008sparse}, and was widely adopted in many applications \cite{elad2006image,mairal2008sparse,huang2007unsupervised,yang2009linear}, we are inspired by the studies showing that general $p$ performs better than special $p$, and ask: is general $L_p$-norm able to outperform $L_0$/$L_1$/$L_2$/$L_{\infty}$-norm on a specific dataset for a specific task? If yes, then which $p$ will be the best (i.e. the case of unknown $p$)? These questions seem not being investigated before, to the best of our knowledge. Especially, \emph{non-sparse coding}, i.e. $p>1$, is clearly different from sparse coding and is the major theme of our study.
\subsection{Outline of Solution}
Analytically solving the $L_p$-norm constrained problems is very difficult as they are not convex optimization. When designing numerical solutions, note that there are two (resp. three) groups of variables in (\ref{pncc-known-p}) (resp. (\ref{pncc-unknown-p})), and it is natural to perform alternate optimization over them iteratively. Previous methods for sparse coding mostly follow this approach, for example in \cite{olshausen1997sparse} and in the well-known K-SVD method \cite{aharon2006rm}. One clear drawback of this approach is the high computational complexity due to the iterative nature. A variant of using early-stopped iterative algorithms to provide fast solution approximations was discussed in \cite{xu2019learning}, built on the overly strong assumption of close-to-orthogonal bases \cite{bansal2018can}. Besides the high complexity, it is not straightforward to extend the sparse coding methods to the cases of non-sparse coding, since they usually leverage the premise of sparse code heuristically but for general $p>1$ it is hard to design such heuristics.

A different approach for sparse coding was proposed by Gregor and LeCun \cite{gregor2010learning}, where an iterative algorithm known as iterative shrinkage and thresholding (ISTA), that was previously used for $L_1$-norm regularized least squares, is unrolled and truncated to construct a multi-layer feed-forward network. The network can be trained end-to-end to act as a regressor from a vector to its corresponding sparse code. Note that truncation helps lower the computational cost of the network than the original algorithm, while training helps compensate the error due to truncation. The trained network known as learned ISTA (LISTA) is then a fast alternative to the original ISTA algorithm. Following the approach of LISTA, several recent works consider the cases of $L_0$-norm \cite{wang2016learning} and $L_{\infty}$-norm \cite{wang2016learning2}, as well as extend other iterative algorithms to their network versions \cite{ross2011learning,wang2016proximal,sun2016deep}. However, LISTA and its following-up works are not applicable for solving the cases of general $p$, because they all refer to the same category of first-order iterative algorithms, i.e. the projected gradient descent (PGD) algorithms. For general $p$, the projection step in PGD is not analytically solvable. In addition, such works have more difficulty in solving the unknown $p$ problem.

Beyond the family of PGD, another first-order iterative algorithm is the Frank-Wolfe algorithm \cite{frank1956algorithm}, which is free of projection. This characteristic inspires us to adapt the Frank-Wolfe algorithm to solve the general $p$ problem. Similar to LISTA, we unroll and truncate the Frank-Wolfe algorithm to construct a network, termed Frank-Wolfe network (F-W Net), which can be trained end-to-end. Although the convergence speed of the original Frank-Wolfe algorithm is slow, we will show that F-W Net can converge faster than not only the original Frank-Wolfe and ISTA algorithms, but also the LISTA.
Moreover, F-W Net has a novel, closed-form and nonlinear computation unit that is parameterized by $p$, and as $p$ varies, that unit displays the behaviors of several classic pooling operators, and can be naturally viewed as a cascade of a generalized normalization and a parametric activation. Due to the fact that $p$ becomes a (hyper) parameter in F-W Net, we can either set $p$ a priori or make $p$ learnable when training F-W Net. Thus, F-W Net has higher learning flexibility than LISTA and training F-W Net gracefully solves the unknown $p$ problem.
\subsection{Our Contributions}
To the best of our knowledge, we are the first to extend the sparse coding problem into $L_p$-norm constrained coding with general $p>1$ and unknown $p$; we are also the first to unroll-and-truncate the Frank-Wolfe algorithm to construct trainable network. Technically, we make the following contributions:
\begin{itemize}
  \item We propose the Frank-Wolfe network (F-W Net), whose architecture is inspired by unrolling and truncating the Frank-Wolfe algorithm for solving $L_p$-norm regularized least squares problem. F-W Net features a novel nonlinear unit that is parameterized by $p$ and termed $pool_p$. The $pool_p$ unit links the conventional pooling, activation, and normalization operations in deep networks. F-W Net is verified to solve non-sparse coding with general known $p>1$ better than the existing iterative algorithms. More importantly, F-W Net solves the non-sparse coding with unknown $p$ at low computational costs.
  \item We propose a convolutional version of F-W Net, which extends the basic F-W Net by adding convolutions and utilizing the $pool_p$ unit point by point across different channels. The convolutional (Conv) F-W Net can be readily applied into image-related tasks.
  \item We evaluate the performance of F-W Net on an extensive range of simulations as well as a handwritten digit recognition task, where F-W Net exhibits strong learning capability. We further apply Conv F-W Net into image denoising and super-resolution, where it also demonstrates impressive effectiveness, flexibility, and robustness.
\end{itemize}

\subsection{Paper Organization}
The remainder of this paper is organized as follows. Section \ref{sec_relatedwork} reviews literatures from four folds: $L_p$-norm constrained coding and its applications, deep structured networks, the original Frank-Wolfe algorithm, and nonlinear units in networks, from which we can see that F-W Net connects and integrates these separate fields. Section \ref{sec_fwnet} formulates F-W Net, with detailed discussions on its motivation, structure, interpretation, and implementation issues. Section \ref{sec_simulation} validates the performance of F-W Net on synthetic data under an extensive range of settings, as well as on a toy problem, handwritten digit recognition with the MNIST dataset. Section \ref{sec_convfwnet} discusses the proposed convolutional version of F-W Net. Section \ref{sec_experiments} then provides experimental results of using Conv F-W Net on image denoising and super-resolution, with comparison to some other CNN models. Section \ref{sec_conclusion} concludes this paper. For reproducible research, our code and pre-trained models have been published online\footnote{\url{https://github.com/sunke123/FW-Net}}.

\section{Related Work}\label{sec_relatedwork}
\subsection{$L_p$-Norm Constrained Coding and Its Applications}
Sparse coding as a representative methodology of the linear representation methods, has been used widely in signal processing and computer vision, such as image denoising, deblurring, image restoration, super-resolution, and image classification \cite{elad2006image, wright2009robust, mairal2008sparse, yang2009linear}. The sparse representation aims to preserve the principle component and reduces the redundancy in the original signal.
From the viewpoint of different norm minimizations used in sparsity constraints, these methods can be roughly categorized into the following groups: 1) $L_0$-norm minimization; 2) $L_p$-norm ($0<p<1$) minimization; 3) $L_1$-norm minimization; and 4) $L_{2,1}$-norm minimization.
In addition, $L_{2}$-norm minimization is extensively used, but it does not lead to sparse solution.

Varieties of dictionary learning methods have been proposed and implemented based on sparse representation. K-SVD \cite{aharon2006rm} seeks an over-complete dictionary from given training samples under the $L_0$-norm constraint, which achieves good performance in image denoising. Wright \emph{et al.} \cite{wright2009robust} proposed a general classification algorithm for face recognition based on $L_1$-norm minimization, which shows if the sparsity can be introduced into the recognition problem properly, the choice of features is not crucial. Krogh \emph{et al.} \cite{krogh1992simple} showed that limiting the growth of weights through $L_2$-norm penalty can improve generalization in a feed-forward neural network. The simple weight decay has been adopted widely in machine learning.
\subsection{Deep Structured Networks}
Deep networks are typically stacked with off-the-shelf building blocks that are jointly trained with simple loss functions. Since many real-world problems involve predicting statistically dependent variables or related tasks, deep structured networks \cite{chen2015learning,schwing2015fully} were proposed to model complex patterns by taking into account such dependencies. Among many efforts, a noticeable portion has been devoted to unrolling the traditional optimization and inference algorithms into their deep end-to-end trainable formats.

Gregor and LeCun \cite{gregor2010learning} first leveraged the idea to construct feed-forward networks as fast trainable regressors to approximate the sparse code solutions, whose idea was expanded by many successors, e.g. \cite{sprechmann2015learning,wang2016learning,wang2016learning2,wang2016learning3,wang2016d3,wang2017doubly,zhang2018ista}. Those works show the benefits of incorporating the problem structure into the design of deep architectures, in terms of both performance and interpretability \cite{wisdom2016interpretable}. A series of works \cite{xin2016maximal,moreau2016understanding,chen2018theoretical,liu2018alista} established the theoretical background of this methodology. Note that many previous works \cite{gregor2010learning,sprechmann2015learning, wang2016learning, xin2016maximal} built their deep architectures based on the \emph{projected gradient descent} type algorithms, e.g. the iterative shrinkage and thresholding algorithm (ISTA). The projection step turned into the nonlinear activation function. Wang \emph{et al.} \cite{wang2016proximal} converted proximal methods to deep networks with continuous output variables. More examples include the message-passing inference machine \cite{ross2011learning}, shrinkage fields \cite{schmidt2014shrinkage}, CRF-RNN \cite{zheng2015conditional}, and ADMM-net \cite{sun2016deep}.

\subsection{Frank-Wolfe Algorithm}
The Frank-Wolfe algorithm \cite{frank1956algorithm}, also known as conditional gradient descent, is one of the simplest and earliest known iterative solver, for the generic constrained convex problem:
\begin{align}\label{eq:problem}
\min_{\mathbf{z}} ~f(\mathbf{z})~~
\text{s.t.} ~\mathbf{z} \in \mathcal{Z}
\end{align}
where $f$ is a convex and continuously differentiable objective function, and $\mathcal{Z}$ is a convex and compact subset of a Hilbert space. At each step, the Frank-Wolfe algorithm first considers the linear approximation of $f(\mathbf{z})$, and then moves towards this linear minimizer that is taken over $\mathcal{Z}$. Section \ref{sec_fwnet} presents a concrete example of applying the Frank-Wolfe algorithm.

The Frank-Wolfe algorithm has lately re-gained popularity due to its promising applicability in handling structural constraints, such as sparsity or low-rank. The Frank-Wolfe algorithm is projection-free: while competing methods such as the projected gradient descent and proximal algorithms need to take a projection step back to the feasible set per iteration, the Frank-Wolfe algorithm only solves a linear problem over the same set in each iteration, and automatically stays in the feasible set. For example, the sparsity regularized problems are commonly relaxed as convex optimization over convex hulls of atomic sets, especially $L_p$-norm constrained domains \cite{chandrasekaran2012convex}, which makes the Frank-Wolfe algorithm easily applicable.
We refer the readers to the comprehensive review in \cite{jaggi2013revisiting,zhang2017randomized} for more details about the algorithm.
\subsection{Nonlinear Units in Networks}
There have been blooming interests in designing novel nonlinear activation functions \cite{agostinelli2014learning}, a few of which have parametric and learnable forms, such as the parametric ReLU \cite{he2015delving}. Among existing deep structured networks, e.g. \cite{gregor2010learning,sprechmann2015learning, wang2016learning}, their activation functions usually took fixed forms (e.g. some variants of ReLU) that reflected the pre-chosen structural priors.
A data-driven scheme is presented in \cite{kamilov2016learning} to learn optimal thresholding functions for ISTA. Their adopted parametric representations led to spline curve-type activation function, which reduced the estimation error compared to using the common (fixed) piece-wise linear functions.

As another major type of nonlinearity in deep networks, pooling was originally introduced as a dimension-reduction tool to aggregate a collection of inputs into low-dimensional outputs \cite{jarrett2009best}. Other than the input-output dimensions, the difference between activation and pooling also lies in that activation is typically applied element wise, while pooling is on groups of hidden units, usually within a spatial neighborhood. It was proposed in \cite{malinowski2013learnable} to learn task-dependent pooling and to adaptively reshape the pooling regions. More learnable pooling strategies are investigated in \cite{lee2016generalizing}, via either mixing two different pooling types or a tree-structured fusion. Gulcehre \emph{et al.} \cite{gulcehre2014learned} introduced the $L_p$ unit that computed a normalized $L_p$ norm over the set of outputs, with the value of $p$ learnable.

In addition, we find our proposed nonlinear unit inherently related to normalization techniques. Jarrett \emph{et al.} \cite{jarrett2009best} demonstrated that a combination of nonlinear activation, pooling, and normalization improved object recognition. Batch normalization (BN) \cite{ioffe2015batch} rescaled the summed inputs of neurons over training batches, and substantially accelerated training. Layer normalization (LN) \cite{ba2016layer} normalized the activations across all activities within a layer. Ren \emph{et al.} \cite{ren2016normalizing} re-exploited the idea of divisive normalization (DN) \cite{lyu2010divisive, balle2015density}, a well-grounded transformation in real neural systems. The authors viewed both BN and LN as special cases of DN, and observed improvements by applying DN on a variety of tasks.

\section{Frank-Wolfe Network}\label{sec_fwnet}
\subsection{Frank-Wolfe Solver for $L_p$-Norm Constrained Least Squares}
We investigate the $L_p$-norm constrained least squares problem as a concrete example to illustrate the construction of F-W Net. The proposed methodology can certainly be extended to more generic problems (\ref{eq:problem}). Let $\mathbf{x} \in \R^n$ denote the input signal, and $\mathbf{z} \in \R^m$ denote the code (a.k.a. representation, feature vector). $D \in \R^{n \times m}$ is the decoding matrix (a.k.a. dictionary, bases). The problem considers $\mathcal{Z}$ to be an $L_p$-norm ball of radius $c$, with $p \ge 1$ to ensure the convexity of $\mathcal{Z}$, and $f(\mathbf{z})$ to be a least-squares function:
\begin{align}\label{lpr}
\mathbf{z}^* = \arg \min_{\mathbf{z}} \frac{1}{2}||\mathbf{x} - D \mathbf{z}||_2^2  ~~\text{s.t.}~~||\mathbf{z}||_p \leq  c.
\end{align}
We initialize $\mathbf{z}^0 \in \mathcal{Z}$. At iteration $t = 0, 1, 2, \dots$, the Frank-Wolfe algorithm iteratively updates two variables $\mathbf{z}^t, \mathbf{s}^t \in \R^m$ to solve (\ref{lpr}):
\begin{align}
\mathbf{s}^t:= & \arg\min_{\mathbf{s} \in \mathcal{Z}} ~ \mathbf{s}^{\top} \nabla f(\mathbf{z}^t)\nonumber\\
\mathbf{z}^{t+1}:=& (1-\gamma^t)\mathbf{z}^t+\gamma^t\mathbf{s}^t, 
\label{fwsolver}
\end{align}
where $\nabla f(\mathbf{z}^t) = D^\top(D \mathbf{z}^t - \mathbf{x})$.  $\gamma^t$ is the step size, which is typically set as $\gamma^t := \frac{2}{t+2}$ or chosen via line search. By H\"older's inequality, the solution of $\mathbf{s}^t$ is:
\begin{align}
\begin{split}
s_i^t = - c \cdot \text{sign}(\nabla_i f(\mathbf{z}^t)) \frac{|\nabla_i f(\mathbf{z}^t)|^{\frac{1}{p-1}}}{(\sum_{j=1}^m|\nabla_j f(\mathbf{z}^t)|^{\frac{p}{p-1}})^{\frac{1}{p}}}, \\
i = 1, 2, ..., m
\label{pool}
\end{split}
\end{align}
where $s_i^t, \nabla_i f(\mathbf{z}^t)$ denote the $i$-th entry of $\mathbf{s}^t, \nabla f(\mathbf{z}^t)$, respectively. It is interesting to examine a few special $p$ values in (\ref{pool}) (ignoring the negative sign for simplicity):
\begin{itemize}
	\item $p =1$, $\mathbf{s}^t$ selects the largest entry in $\nabla f(\mathbf{z}^t)$, while setting all other entries to zero\footnote{This reminds us of the well-known matching pursuit algorithm. We noted that a recent work \cite{locatello2017unified} has revealed a unified view between the Frank-Wolfe algorithm and the matching pursuit algorithm.}.
	\item $p =2$, $\mathbf{s}^t$ re-scales $\nabla f(\mathbf{z}^t)$ by its root mean square.
	\item $p = \infty$,  
	all entries of $\mathbf{s}^t$ have the equal magnitude $c$.
\end{itemize}
The three special cases easily remind the behaviors of \textit{max~pooling}, \textit{root-mean-square~pooling} and \textit{average~pooling} \cite{jarrett2009best, yang2009linear}, although the input and output are both $\R^m$ and no dimensionality reduction is performed. For general $p \in [1, \infty)$, it is expected to exhibit more varied behaviors that can be interpretable from a pooling perspective. We thus denote by the function $\R^m \rightarrow \R^m$: $\mathbf{s}^t$ = $pool_p (\nabla f(\mathbf{z}^t))$, to illustrate the operation (\ref{pool}) associated with a specific $p$.
\subsection{Constructing the Network}
\begin{figure*}
	\centering
	\includegraphics[width=\textwidth]{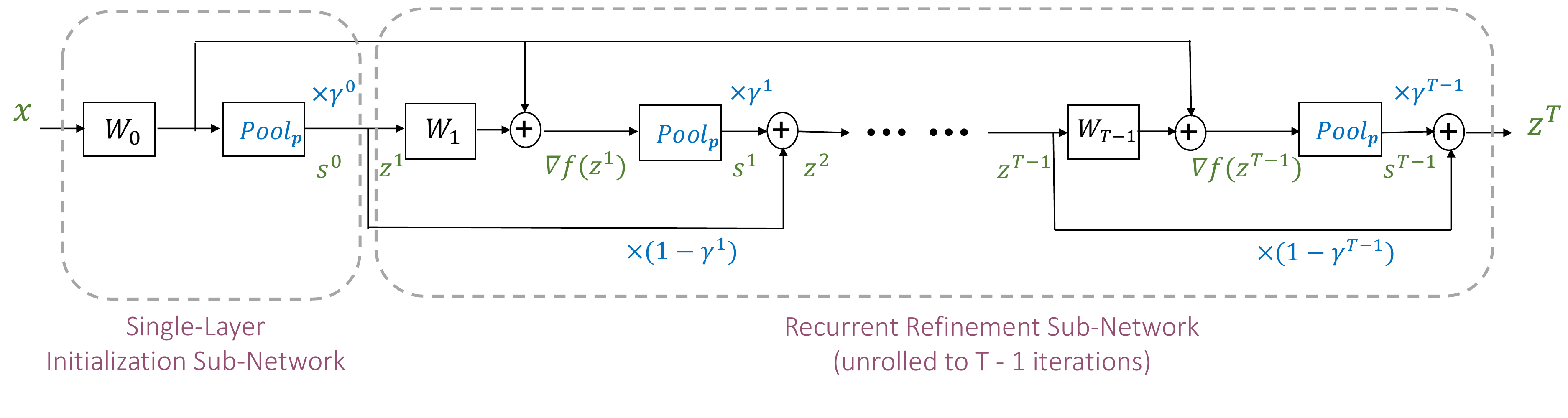}
\caption{(Please view in color) $T$-layer Frank-Wolfe network, consisting of two parts: (1) a single-layer initialization sub-network; and (2) a recurrent refinement sub-network, unrolled to $T-1$ iterations. We use the \emph{Green} notations to remind to which variable in (\ref{fwsolver}) the current intermediate output is corresponding. The layer-wise weights are denoted in \emph{Black}, and the learnable units/hyper-parameters are in \emph{Blue}.}\label{fig_fwnet}
\end{figure*}
Following the well received unrolling-then-truncating methodology, e.g. \cite{gregor2010learning}, we represent the Frank-Wolfe solver for (\ref{lpr}) that runs finite $T$ iterations ($t = 0, 1, \dots, T -1$), as a multi-layer feed-forward neural network. Fig. \ref{fig_fwnet} depicts the resulting architecture, named the Frank-Wolfe Network (F-W Net). As a custom, we set $\mathbf{z}^0 = \mathbf{0}$, which is an interior point of $\mathcal{Z}$ for any $p \ge 1$. The $T$ layer-wise weights $W_t$'s can be analytically constructed. Specifically, note that $\nabla f(\mathbf{z}^t) = D^\top(D \mathbf{z}^t - \mathbf{x})= D^\top D \mathbf{z}^t - D^\top\mathbf{x}$, if we define
\begin{align}
W_0 =  -D^\top; W_t =  D^\top D, t = 1, 2, ...., T - 1.
\label{weight}
\end{align}
Then we have $\nabla f(\mathbf{z}^t)=W_t\mathbf{z}^t+W_0\mathbf{x}$, as depicted in Fig. \ref{fig_fwnet}.
Given the pre-chosen hyper-parameters, and without any further tuning, F-W Net outputs a $T$-iteration approximation $\mathbf{z}^T$ of the exact solution $\mathbf{z}$ to (\ref{lpr}). As marked out, the intermediate outputs of F-W Net in Fig. \ref{fig_fwnet} are aligned with the iterative variables in the original Frank-Wolfe solver (\ref{fwsolver}). Note that the two-variable update scheme in (\ref{fwsolver}) naturally leads to two groups of ``skip connections,'' which might be reminiscent of the ResNet \cite{he2016deep}.

As many existing works did \cite{gregor2010learning,sprechmann2015learning,wang2016learning}, the unrolled and truncated network in Fig. \ref{fig_fwnet} could be alternatively treated as a trainable regressor to predict the exact $\mathbf{z}$ from $\mathbf{x}$. We further view F-W Net as composed from two sub-networks: (1) a single-layer initialization sub-network, consisting of a linear layer $W_0$ and a subsequent $pool_p$ operator. It provides a rough estimate of the solution: $\mathbf{z}_1 = \gamma^0\mathbf{s}_0 = \gamma^0 pool_p (W_0 \mathbf{x})$, which appears similar to typical sparse coding that often initializes $\mathbf{z}$ with a thresholded $D^T\mathbf{x}$ \cite{wang2016sparse}. Note that $\mathbf{z}_1$ has a direct shortcut path to the output, with the multiplicative weights $\prod_{t=1}^{T-1} (1- \gamma^t)$; (2) a recurrent refinement sub-network, that is unrolled to repeat the layer-wise transform for $T-1$ times to gradually refine the solution.

F-W Net is designed for large learning flexibility, by fitting almost all its parameters and hyper-parameters from training data:
\begin{itemize}
	\item \textbf{Layer-wise weights.} Eq. (\ref{weight}) can be used to initialize the weights, but during training, $W_0$ and $W_t$ are untied with $D$ and viewed as conventional fully-connected layers (without biases). All weights are learned with back-propagation from end to end. In some cases (e.g. Table \ref{tab_mnist_p}), we find that sharing the $W_t$'s (with $t = 1, 2, ...., T - 1$) is a choice worthy of consideration: it effectively reduces the actual number of parameters and makes the training to converge fast. But not sharing the weights is always helpful to improve the performance, as observed in our experiments. Thus, the weights are not shared unless otherwise noted. In addition, the relation between $W_0$ and $W_t$ (with $t = 1, 2, ...., T - 1$), as in (\ref{weight}), is not enforced during training. $W_0$ is treated as a simple fully-connected layer without additional constraint.
	\item \textbf{Hyper-parameters.} $p$ and $\gamma^t$ were given or pre-chosen in (\ref{fwsolver}). In F-W Net, they can be either pre-chosen or made learnable. For learnable hyper-parameters, we also compute the gradients w.r.t. $p$ and $\gamma^t$, and update them via back-propagation. We adopt the same $p$ throughout F-W Net as $p$ implies the structural prior. $\gamma^t$ is set to be independent per layer. Learning $p$ and $\gamma^t$ also adaptively compensates for the truncation effect \cite{kamilov2016learning} of iterative algorithms. In addition, learning $p$ gracefully solves the unknown $p$ problem (\ref{pncc-unknown-p}).
\end{itemize}
F-W Net can further be jointly tuned with a task-specific loss function, e.g. the softmax loss for classification, or the mean-squared-error loss and/or a semantic guidance loss \cite{liu2018image} for denoising,
in the form of an end-to-end network.
\subsection{Implementing the Network}
\subsubsection{Reformulating $pool_p$ as normalization plus neuron}
A closer inspection on the structure of the $pool_p$ function (\ref{pool}) leads to a two-step decomposition (let $\mathbf{u}^t = \nabla f(\mathbf{z}^t)$ for simplicity):
\begin{align}
\mathbf{y}^{t}=& \frac{\mathbf{u}^t}{||\mathbf{u}^t||_{\frac{p}{p-1}}} ~~~~~~~~~~~~~~\textit{// Step 1:} \, p\textit{-conjugate normalization}\nonumber\\
s_i^{t} = & \, - c \cdot \text{sign}(y_i^t) \cdot |y_i^t|^{\frac{1}{p-1}} ~~ \textit{// Step 2:} \, p\textit{-exponential neuron}
\label{decom}
\end{align}
Step 1 performs a normalization step under the $\frac{p}{p-1}$-norm. Let $q= \frac{p}{p-1}$, which happens to be the H\"older conjugate of $p$: we thus call this step $p$-conjugate normalization. It coincides with a simplified form of DN \cite{ren2016normalizing}, by setting all summation and suppression weights of DN to 1.

Step 2 takes the form of an exponential-type and non-saturated element-wise activation function \cite{clevert2015fast}, and is a learnable activation parameterized by $p$ \cite{agostinelli2014learning}. As the output range of Step 1 is $[-1,1]$, the exponent displays suppression effect when $p \in (1, 2)$, and amplifies entries when $p \in (2, \infty)$.

While the decomposition of  (\ref{pool}) is not unique, we carefully choose (\ref{decom}) due to its effectiveness and numerical stability. As a counterexample, if we adopt another more straightforward decomposition of (\ref{pool}):
\begin{align}
y_i^t =  \text{sign}(u_i^t) \cdot |u_i^t|^{\frac{1}{p-1}}; ~~ \mathbf{s}^{t}=  - \frac{c \cdot \mathbf{y}^t}{||\mathbf{y}^t||_p},
\label{decom2}
\end{align}
then, large $|u_i^t|$ values ($>1$) will be boosted by the power of $\frac{1}{p-1}$ when $p \in (1, 2)$, and the second step may run the risk of explosion when $p \rightarrow 1$, in both feed-forward and back-propagation. In contrast, (\ref{decom}) first squashes each entry into $[-1,1]$ (Step 1) before feeding into the exponential neuron (Step 2), resolving the numerical issue well in practice.

The observation (\ref{decom}), called ``\textit{$pool_p$ =  normalization + neuron}'' for brevity, provides a new interesting insight into the connection between neuron, pooling and normalization, the three major types of nonlinear units in deep networks that were once considered separately. By splitting $pool_p$ into two modules sharing the parameter $p$, the back-propagation computation is also greatly simplified, as directly computing the gradient of $pool_p$ w.r.t. $p$ can be quite involved, with more potential numerical problems.
\subsubsection{Network initialization and training}
The training of F-W Net evidently benefits from high-quality initializations. Although $W_t$ and $D$ are disentangled, $W_t$ can be well initialized from the given or estimated $D$\footnote{For example, when $D$ is not given, we can estimate $D$ using K-SVD \cite{aharon2006rm} to initialize $W$.} via (\ref{weight}). $p$ is typically initialized with a large scalar, but its learning process is found to be quite insensitive to initialization. As observed in our experiments, no matter what $p$ is initialized as, it will converge stably and smoothly to almost the same value\footnote{We are aware of the option to re-parameterize $p$ to ensure $p \ge 1$ \cite{gulcehre2014learned}. We have not implemented it in our experiments, since we never encountered $p < 1$ during learning. The re-parameterization trick can be done if necessary.}. $\gamma^t$ is initialized with the rule $\frac{2}{t+2}$, and is re-parametrized using a sigmoid function to enforce $[0,1]$ range during training.
$c$ is the only hyper-parameter that needs to be manually chosen and fed into F-W Net. In experiments, we find that a good $c$ choice could accelerate the convergence.

In the original $p$-NCLS, $||D||_2 = 1$ is assumed to avoid the scale ambiguity, and is commonly enforced in dictionary learning \cite{aharon2006rm}. Similarly, to restrain the magnitude growth of $W$, we adopt the $L_2$-norm weight decay regularizer, with the default coefficient $2 \times 10^{-4}$ in experiments.
\subsection{Interpretation of Frank-Wolfe Network as LSTM}\label{lstm}
The ISTA algorithm \cite{wisdom2016interpretable} has been interpreted as a stack of plain Recurrent Neural Networks (RNNs). We hereby show that F-W Net can be potentially viewed as a special case of the Long Short-Term Memory (LSTM) architecture \cite{hochreiter1997long}, which incorporates a gating mechanism \cite{chung2014empirical} and may thus capture long-term dependencies better than plain RNNs when $T$ is large. Although we include no such experiment in this paper, we are interested in applying F-W Net as LSTM to model sequential data in future work.

Let us think $\mathbf{x}$ as a constant \textit{input}, and $\mathbf{z}^{t}$ is the \textit{previous hidden state} of the $t$-th step ($t = 1, ... , T-1$). The current \textit{candidate hidden state} is computed by $\mathbf{s}^t = pool_p (W_0 \mathbf{x} + W_t \mathbf{z}^{t})$, where the sophisticated $pool_p$ function replaces the common \texttt{tanh} to be the activation function. $\gamma_t$ and $(1-\gamma_t)$ each take the role of the \textit{input gate} and \textit{forget gate}, that control how much of the newly computed and previous state will go through, respectively. $\gamma^t\mathbf{s}^t + (1-\gamma^t)\mathbf{z}^t$ constitutes the \textit{internal memory} of the current unit. Eventually, with a simple \textit{output gate} equal to 1, the \textit{new hidden state} is updated to $\mathbf{z}^{t+1}$.
\section{Evaluation of Frank-Wolfe Network}\label{sec_simulation}
\subsection{Simulations}
We generate synthetic data in the following steps. First we generate a random vector $\mathbf{y}\in\R^m$ and then we project it to the $L_p$-ball of radius $c$. The projection is achieved by solving the problem: $\mathbf{z}=\arg\min\frac{1}{2}\|\mathbf{y}-\mathbf{z}\|^2_2 \mbox{ s.t. }\|\mathbf{z}\|_p\leq c$, using the original F-W algorithm until convergence. This $p$ will be termed \emph{real} $p$ in the following. We then create a random matrix $D \in \R^{n \times m}$, and achieve $\mathbf{x}: = D\mathbf{z} + \mathbf{e}$, where $\mathbf{e} \in \R^{n}$ is additive white Gaussian noise with variance $\sigma^2=0.01$. We use the default values $m = 100$, $n = 50$, $c = 5$, and generate 15,000 samples for training. A testing set of 1,000 samples are generated separately in the identical manner. Our goal is to train a F-W Net to predict/regress $\mathbf{z}$ from the given observation $\mathbf{x}$. The performance is measured by mean-squared-error (MSE) between predicted and ground-truth $\z$. All network models are implemented with Caffe \cite{jia2014caffe} and trained by using MSE between ground-truth and network-output vectors as loss.

We first choose real $\bm{p=1.3}$, vary $T = 2, 4, 6, 8$, and compare the following methods:
\begin{itemize}
	\item \textit{CVX}: solving the problem using the ground-truth $D$ and real $p$. The CVX package \cite{grant2008cvx} is employed to solve this convex problem. No training process is involved here.
	\item \textit{Original F-W}: running the original Frank-Wolfe algorithm for $T$ iterations, using the ground-truth $D$ and real $p$ and fixing $\gamma^t$ = $\frac{2}{t+2}$. No training process is involved here.
    \item \textit{MLP}: replacing the $pool_p$ in F-W Net with ReLU, having a feed-forward network of $T$ fully-connected layers, which has the same number of parameters with F-W Net (except $p$).
	\item \textit{F-W Net}: the proposed network that jointly learns $W_t$, $p$, and $\gamma^t,t = 0, 1, ..., T-1$.
	\item \textit{F-W fixed \{$p$, $\gamma$\}}: fixing $p = 1.3$ and $\gamma^t$ = $\frac{2}{t+2}$, learning $W_t$ in F-W Net.
	\item \textit{F-W fixed $\gamma$}: fixing $\gamma^t$ = $\frac{2}{t+2}$, learning $W_t$ and $p$ in F-W Net.
	\item \textit{F-W fixed \{wrong $p$, $\gamma$\}}: fixing $p = 1$ (i.e. an incorrect structural prior) and $\gamma^t$ = $\frac{2}{t+2}$, learning $W_t$ in F-W Net.
\end{itemize}

\begin{figure}
	\centering
	\subfloat[Training Error - Epoch]{
		\includegraphics[width=0.9\linewidth]{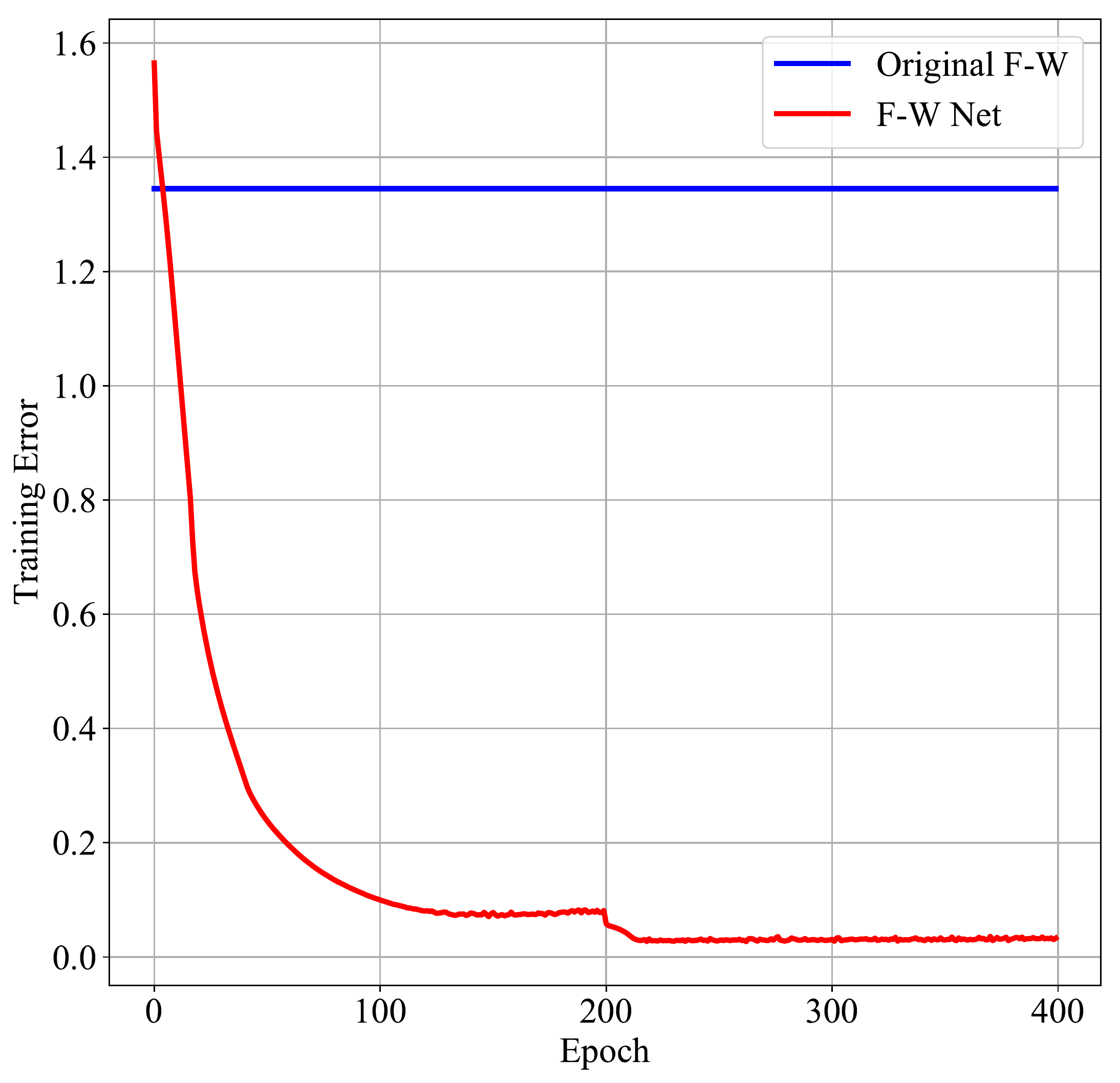}
	}\\
    \subfloat[Testing Error - $T$]{
		\includegraphics[width=0.9\linewidth]{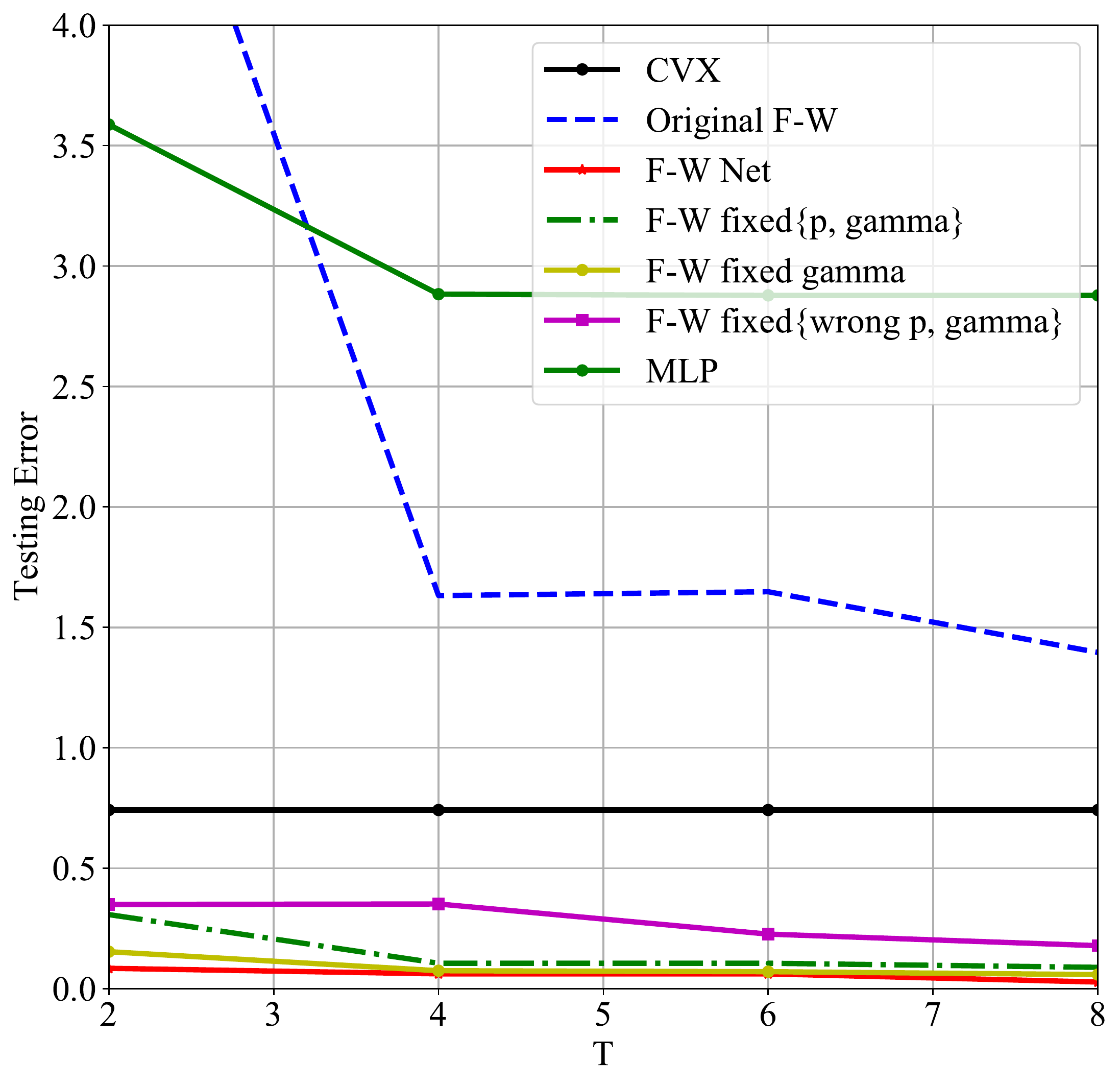}
	}
	\caption{(a) The plot of average training error per sample w.r.t. the epochs at $T=8$. Result of the \textit{Original F-W} is also plotted; (b) Average test error per sample comparison on the testing set, at different $T$'s.}
	\label{fig_fwnet_diffT}
\end{figure}

Fig. \ref{fig_fwnet_diffT} (a) depicts the training curve of F-W net at $T = 8$. We also run the \textit{Original F-W} (for 8 iterations) for comparison. After a few epochs, F-W Net is capable in achieving significantly smaller error than the Original F-W, and converges stably. It shows the advantage of training a network to compensate for the truncation error caused by limited iterations. It is noteworthy that F-W Net has the ability to adjust the dictionary $D$, prior $p$ and step size $\gamma^t$ at the same time, all of which are learned from training data. Comparing the test error of F-W Net and Original F-W at different $T$'s (Fig. \ref{fig_fwnet_diffT} (b)), F-W Net shows the superiority of flexibility, especially when $T$ is small or even the $p$ is wrong. F-W Net learns to find a group parameters ($D$, $p$ and $\gamma^t$) during the training process, so that these parameters coordinate each other to lead to better performance.

Fig. \ref{fig_fwnet_diffT} (b) compares the testing performance of different methods at different $T$'s. F-W Net with learnable parameters achieves the best performance. The \textit{Original F-W} performs poorly at $T = 2$, and its error barely decreases as $T$ increases to 4 and 8, since the original F-W algorithm is known to converge slowly in practice. CVX does not perform well as this problem is quite difficult ($p = 1.3$). Though having the same number of parameters, MLP performs not well, which indicates that this synthetic task is not trivial. Especially, the original Frank-Wolfe algorithm also significantly outperforms MLP after 4 iterations.

Furthermore, we reveal the effect of each component ($D$, $p$ and $\gamma^t$). Firstly, we fix $p$ and $\gamma^t$, i.e. we learn the $W_t$'s compared to the original F-W algorithm. By observing the two curves of testing performance, F-W Net improves the performance with learnable $W_t$'s which coordinate the fixed $p$ and $\gamma^t$, and achieves better approximation of the target $\mathbf{z}$ in a few steps. Then, we let $p$ be learnable and only fix $\gamma^t$, which is slightly superior to \textit{F-W fixed \{$p$, $\gamma$\}}. F-W Net also maintains a smaller, but consistent margin over \textit{F-W fixed $\gamma$}. Those three comparisons confirm that except for $W_t$, learning $p$ and $\gamma$ are both useful. Finally, we give the wrong $p$ to measure the influence on F-W Net. It is noteworthy that \textit{F-W fixed \{wrong $p$, $\gamma$\}} suffers from much larger error than other F-W networks. That demonstrates the huge damage that an incorrect or inaccurate structural prior can cause to the learning process. But, as mentioned before, F-W Net has the high flexibility to adjust the other parameters. Even though under the wrong $p$ condition, F-W Net still outperforms the original F-W algorithm in a few steps.

\begin{figure}
	\centering
	\subfloat[$p$ - Epoch]{
		\includegraphics[width=0.47\linewidth]{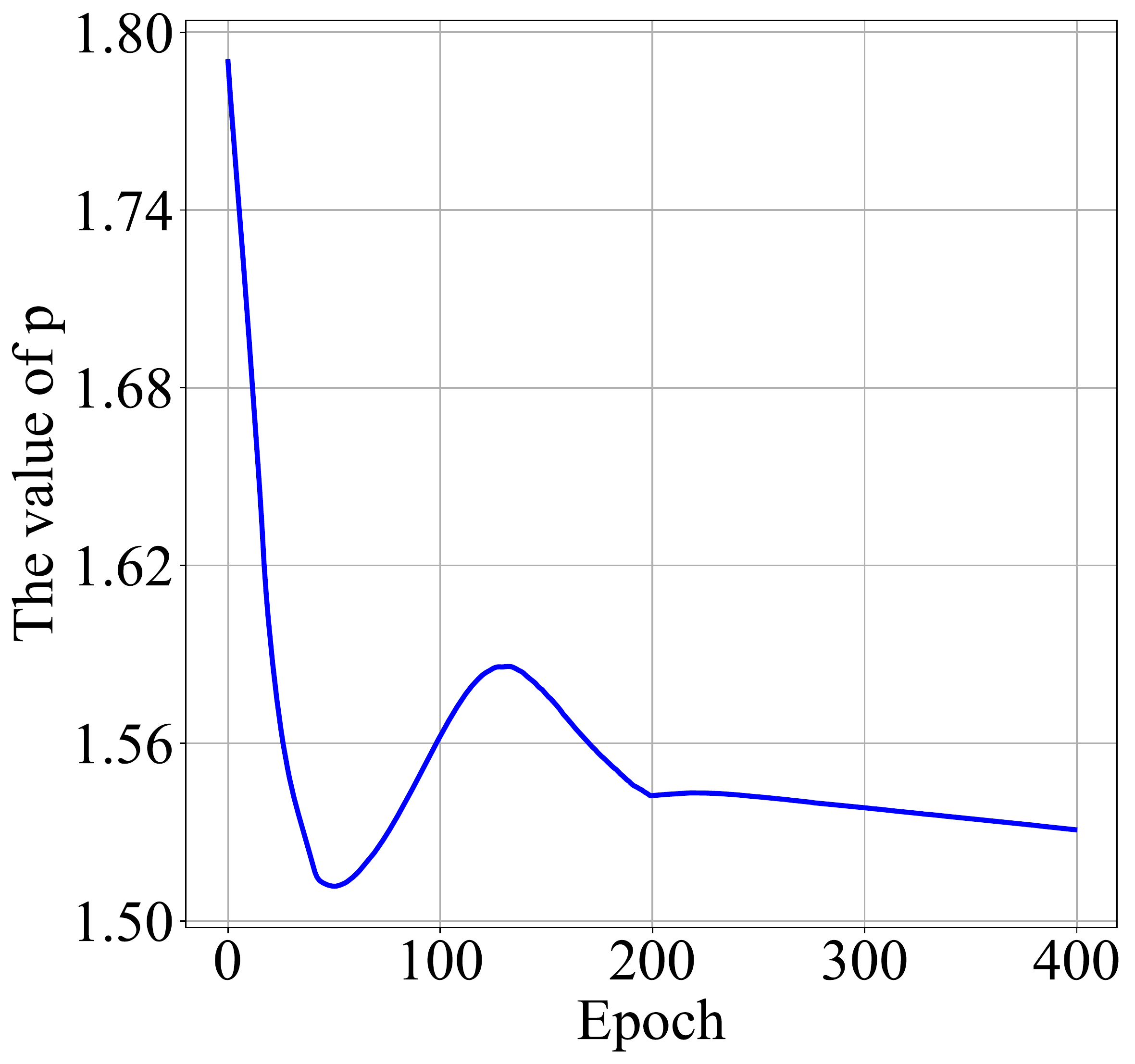}
	}
\subfloat[$\gamma^t$ - $t$]{
		\includegraphics[width=0.47\linewidth]{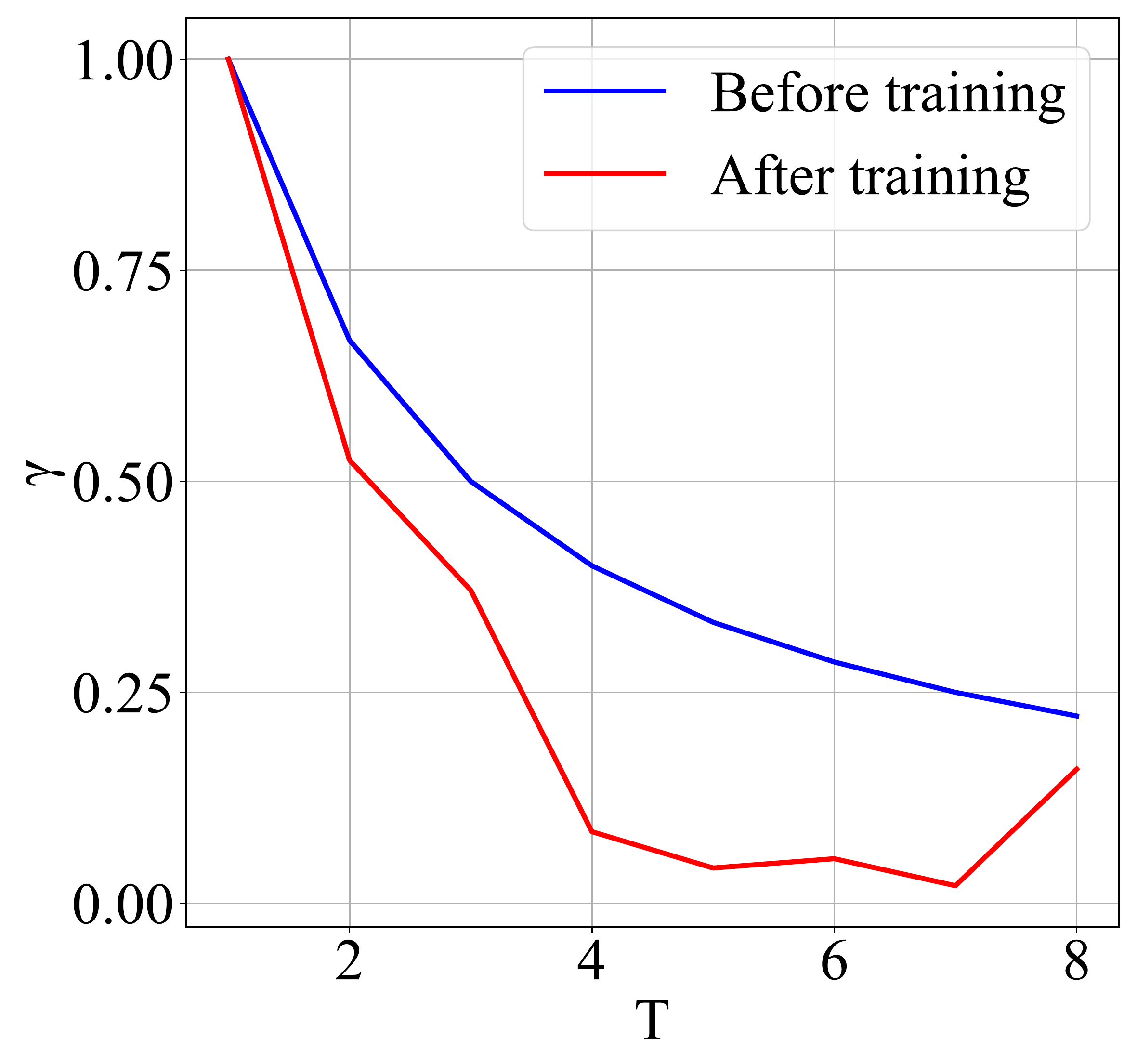}
	}
\\
	\subfloat[$p$ - $T$]{
		\includegraphics[width=0.47\linewidth]{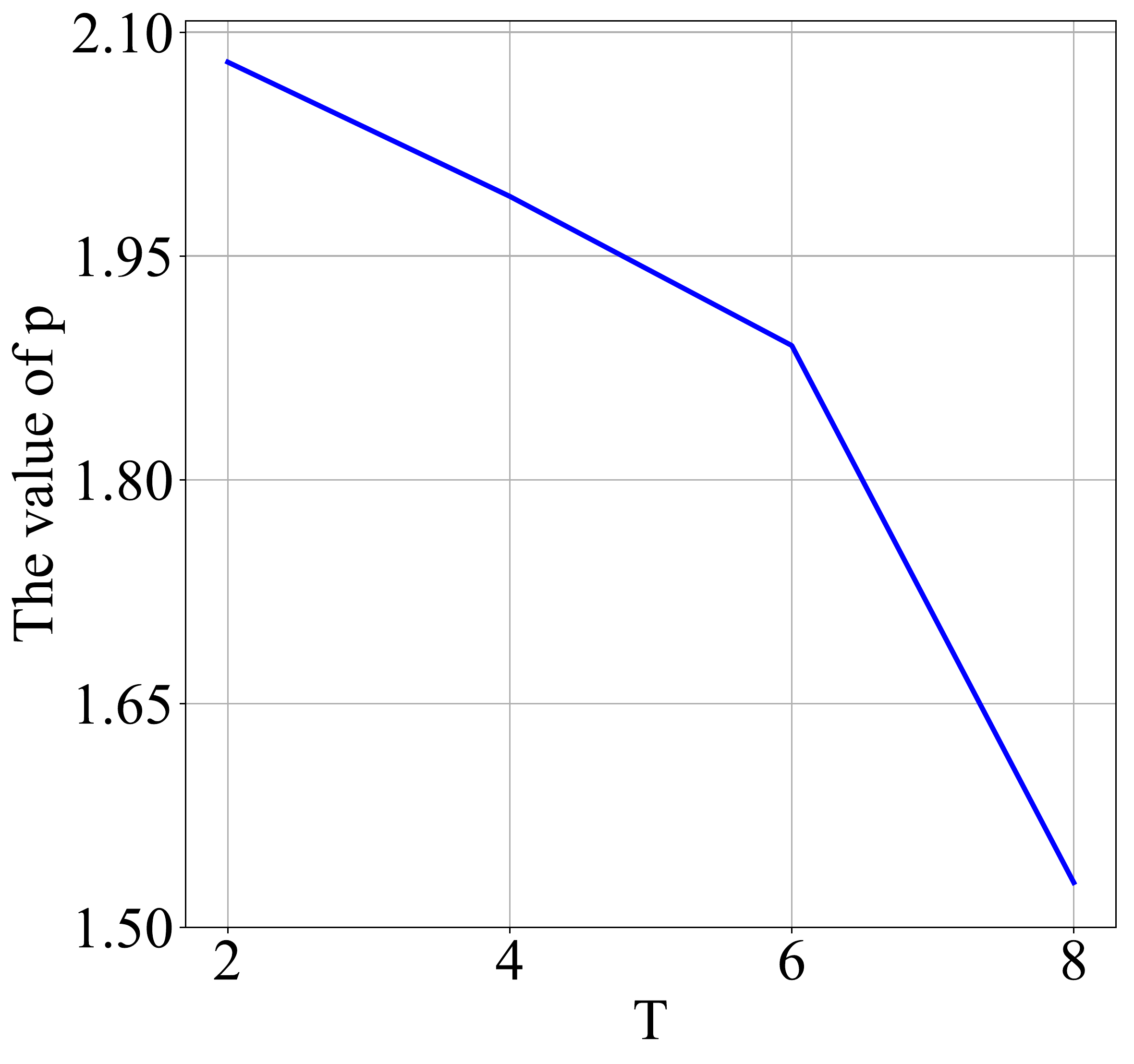}
	}
\subfloat[Learned $p$ - Real $p$]{
		\includegraphics[width=0.47\linewidth]{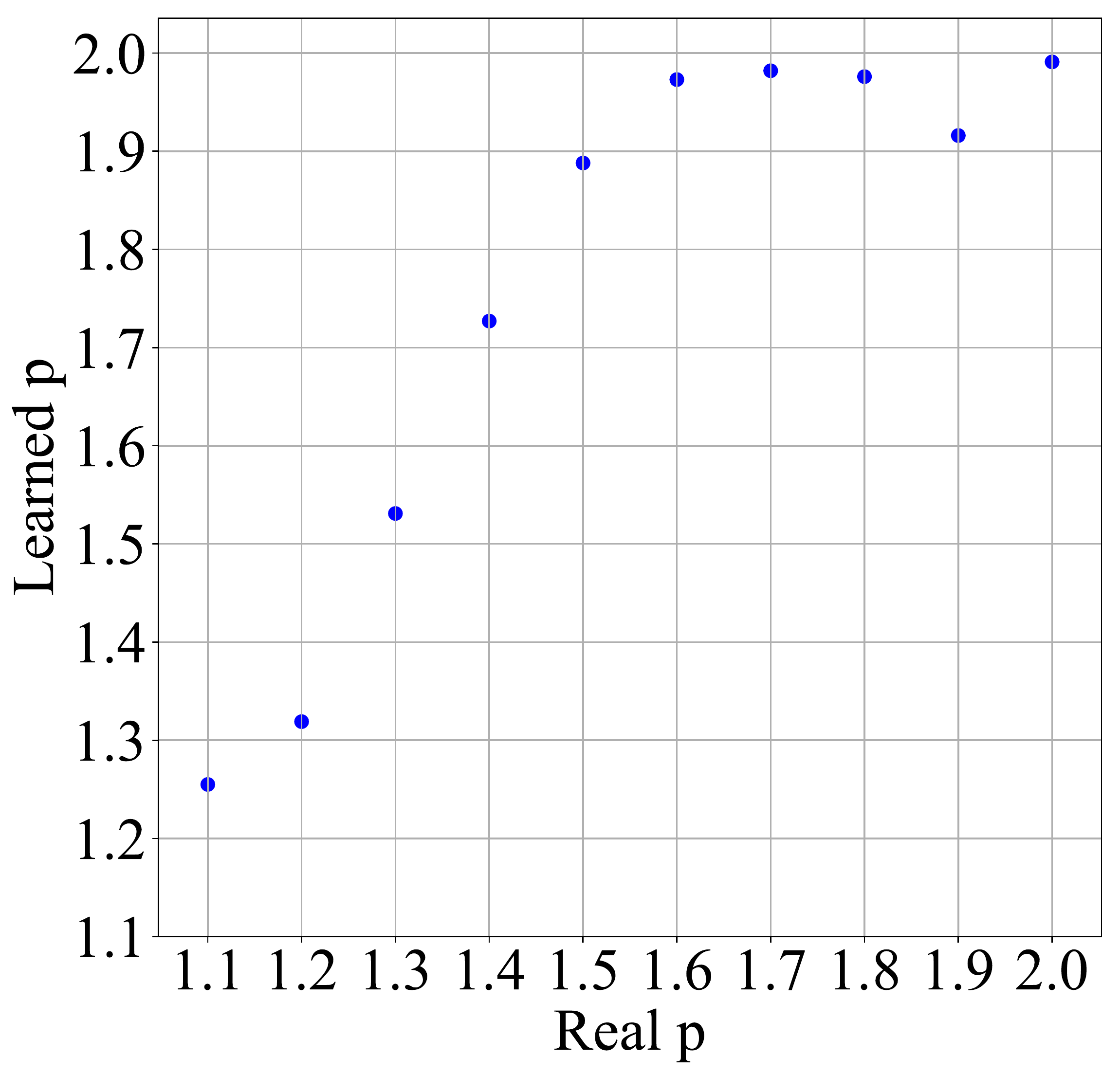}
	}
	\caption{(a) The plot of $p$ during training w.r.t. the epochs at $T=8$ (real $p=1.3$); (b) The plot of $\gamma^t$ values before and after training at $T=8$ (real $p=1.3$); (c) The plot of learned $p$ at different $T$'s (real $p=1.3$); (d) The plot of learned $p$ at $T=8$, when the real $p$ value is from 1.1 to 2.0.}\label{fig_fwnet_learnt}
\end{figure}

Fig. \ref{fig_fwnet_learnt} inspects the learning of $p$ and $\gamma^t$. As seen from Fig. \ref{fig_fwnet_learnt} (a), the $p$ value fluctuates in the middle of training, but ends up converging stably (initial $p=2$ i.e. $L_2$-norm). In Fig. \ref{fig_fwnet_learnt} (c), as $T$ goes up, the learned $p$ by F-W net approaches the real $p = 1.3$ gradually. This phenomenon can be interpreted by that the original F-W algorithm cannot solve the problem well in only a few steps, and thus F-W Net adaptively controls each component through learning them from training data to approximate the distribution of $\mathbf{z}$ as much as possible. To understand why the learned $p$ is usually larger than the real $p$, we may intuitively think that $pool_p$ will ``compress'' the input more heavily as $p$ gets down closer to 1. To predict $\mathbf{z}$, while the original Frank-Wolfe algorithm may run many iterations, F-W Net has to achieve within a much smaller, fixed number of iterations. Thus, each $pool_p$ has to let more energy ``pass through,'' and the learning result has larger $p$. Fig. \ref{fig_fwnet_learnt} (b) observes the change of $\gamma^t$ before and after training, at $T = 8$. While $\gamma^0$ remains almost unchanged, $\gamma^t$ ($t \ge 1$) all decreases. As a possible reason, F-W Net might try to compensate the truncation error by raising the weight of the initialization $\mathbf{z}^1$ in the final output $\mathbf{z}^T$.

\begin{table*}
	\centering
  \caption{Average error per testing sample at different $T$'s. The real $p$ is 1.}\label{tab_fwnet_p1}
		\begin{tabular}{ c | c | ccc | ccc | ccc}
			\hline
			\multirow{2}{*}{} & \multirow{2}{*}{CVX}  & \multicolumn{3}{c|}{F-W Net} & \multicolumn{3}{c|}{F-W Net with fixed $p$} & \multicolumn{3}{c}{LISTA} \\
			\cline{3-11}
			&  & $T = 2$ & $T = 4$ & $T = 6$ & $T = 2$ & $T = 4$ & $T = 6$ & $T = 2$ & $T = 4$ & $T = 6$ \\
			\hline
			MSE & 0.0036 & 0.5641 & 0.3480  & 0.1961 & 1.2076 & 0.8604 & 0.7358 & 0.4157 & 0.3481 & 0.3053 \\
			\hline
			$p$ & 1 (fixed) & 1.360 & 1.254 & 1.222 & 1 (fixed) & 1 (fixed) & 1 (fixed) & 1 (fixed) & 1 (fixed) & 1 (fixed) \\
			\hline
		\end{tabular}
\end{table*}

We then look into $\bm{p=1}$ case, and re-generate synthetic data. We compare three methods as defined before: \textit{CVX},  \textit{F-W Net}, and \textit{F-W fixed $p$}. In addition, we add LISTA \cite{gregor2010learning} into comparison, because LISTA is designed for $p = 1$. The depth, layer-wise dimensions, and weight-sharing of LISTA are configured identically to F-W Net. Note that LISTA is dedicated to the $L_1$ case and cannot be easily extended for general $L_p$ cases. We re-tune all the parameters to get the best performance with LISTA to ensure a fair comparison.
Table \ref{tab_fwnet_p1} compares their results at $T = 2, 4, 6$. The CVX is able to solve $L_1$ problems to a much better accuracy than the case of $p = 1.3$, and F-W Net still outperforms \textit{F-W fixed $p$}. More interesting is the comparison between F-W Net and LISTA: F-W Net is outperformed by LISTA at $T = 2$,  then reaches a draw at $T = 4$, and eventually outperforms LISTA by a large margin at $T = 6$. Increasing $T$ demonstrates a more substantial boost on the performance of F-W Net than that of LISTA, which can be interpreted as we have discussed in Section \ref{lstm} that F-W Net is a special LSTM, but LISTA is a vanilla RNN \cite{wisdom2016interpretable}.
Note that the real $p$ is 1 (corresponding to sparse), but the learned $p$ in F-W Net is larger than 1 (corresponding to non-sparse). The success of F-W Net does not imply that the problem itself is a non-sparse coding one. This is also true for all the following experiments.

We also simulate with other real $p$ values.
Fig. \ref{fig_fwnet_learnt} (d) shows the learned $p$ values with respect to different real $p$'s. When the real $p$ approaches $1$ or $2$, F-W Net is able to estimate the value of $p$ more accurately, probably because the convex problems with $L_1$-norm and $L_2$-norm minimization can be solved more easily.
\subsection{Handwritten Digit Recognition}\label{sec_mnist}
Similar to what has been done in \cite{gregor2010learning}, we adopt F-W Net as a feature extractor and then use logistic regression to classify the features for the task of handwritten digit recognition. We use the MNIST dataset \cite{lecun1998gradient} to experiment. We design the following procedure to pre-train the F-W Net as a feature extractor.
The original images are dimensionality reduced to 128-dim by PCA for input to F-W Net. Then we further perform PCA on each digit separately to reduce the dimension to 15. We construct a 150-dim sparse code for each image, whose 150 dimensions are divided into 10 groups to correspond to 10 digits, only 15 dimensions of which are filled by the corresponding PCA result, whereas the other dimensions are all zero. This sparse code is regarded as the ``ground-truth'' for F-W Net in training. Accordingly, the transformation matrices in the second-step PCA are concatenated to serve as $D\in\R^{128\times 150}$, which is used to initialize the fully-connected layers in F-W Net according to (\ref{weight}).
In this experiment, we use stochastic gradient descent (SGD), a momentum of 0.9 and a mini-batch size of 100. The F-W Net is pre-trained for 200 epochs, and the initial learning rate is 0.1, decayed to 0.01 at 100 epochs. Then the pre-trained F-W Net is augmented with a fully-connected layer with softmax that is randomly initialized, resulting a network that can be trained end-to-end for classification. We observe that the performance benefits from joint training marginally but consistently.

If we formulate this feature extraction problem as $L_p$-norm constrained, then the $p$ is unknown, and most of previous methods adopt $L_1$-norm or $L_2$-norm. Different from them, F-W Net tries to attain a $p$ from training data, which suits the real data better. The results are shown in Table \ref{tab_mnist}, comparing F-W Net with simple feed-forward fully-connected networks (MLP) \cite{lecun1998gradient} and LCoD \cite{gregor2010learning}. F-W Net achieves lower error rate than the others. Especially, it takes 50 iterations for LCoD to achieve an error rate of 1.39, but only 3 layers for F-W Net to achieve 1.34, where the numbers of parameters are similar. Moreover, with the increasing number of layers, F-W Net makes a continuous improvement, which is consistent with the observation in the simulation.

Table \ref{tab_mnist_p} provides the results of different initializations of the hyper-parameter $p$, showing that F-W Net is insensitive to the initialization of $p$ and can converge to the learned $p\approx 1.6$. However, fixing the $p$ value is not good for F-W Net even fixing to the finally learned $p=1.6$. This is interesting as it seems the learnable $p$ provides advantages not only for the final trained model but also for training itself. An adjustable $p$ value may suit for the evolving parameters during training F-W Net, which we plan to study further.
\begin{table}
\centering
\caption{Results on the MNIST test set.}\label{tab_mnist}
	\begin{tabular}{l|c|c}
		\hline
		& \# Params& Error rate (\%)\\
		\hline
		3-layer: 300+100 \cite{lecun1998gradient}  & 266,200 & 3.05 \\
		3-layer: 500+150 & 468,500 & 2.95 \\
		3-layer: 500+300  & 545,000 & 1.53 \\
		\hline
		1-iter LCoD \cite{gregor2010learning} & 65,536  & 1.60 \\
		5-iter LCoD& 65,536  & 1.47 \\
		50-iter LCoD & 65,536 & 1.39 \\
		\hline
		2-layer F-W Net & 43,350  & 2.20 \\
		3-layer F-W Net & 65,850  & 1.34 \\
		4-layer F-W Net & 88,200  & \textbf{1.25} \\
		\hline
	\end{tabular}
\end{table}
\begin{table}
	\centering
  \caption{Results on the MNIST test set of F-W Nets with fixed $p$ and learnable $p$. The $W_t$ parameters are shared across $t=1,\dots,T-1$ in F-W Net.}\label{tab_mnist_p}

	\begin{tabular}{c|cc|cc}
	\hline
  \multirow{2}{*}{Initial $p$}
	&\multicolumn{2}{c|}{Learnable $p$} & \multicolumn{2}{c}{Fixed $p$} \\
	\cline{2-5}
	& Accuracy (\%) & $p$ & Accuracy (\%) & $p$ \\
	\hline
	1.1  & 98.51 & 1.601 & 97.35 &1.1 \\
	1.3  & 98.45 & 1.600 & 97.70 &1.3 \\
	1.5  & 98.43 & 1.595 & 98.30 &1.5 \\
	1.6  & 98.66 & 1.614 & 98.26 &1.6 \\
	1.8  & 98.38 & 1.601 & 97.90 &1.8 \\
	2.0  & 98.50 & 1.585 & 97.78 &2.0 \\
	\hline
	\end{tabular}
\end{table}
\section{Convolutional Frank-Wolfe Network}\label{sec_convfwnet}
Previous similar works \cite{gregor2010learning,sprechmann2015learning, wang2016learning} mostly result in fully-connected networks, as they are unrolled and truncated from linear sparse coding models. Nonetheless, fully-connected networks are less effective than convolutional neural networks (CNNs) when tackling structured multi-dimensional signal such as images. A natural idea to extend this type of works to convolutional cases seems to be convolutional sparse coding \cite{bristow2013fast}, which also admits iterative solvers. However, the re-formulation will be inefficient both memory-wise and computation-wise, as discussed in \cite{sreter2018learned}.

We therefore seek a simpler procedure to build the convolutional F-W Net: in Fig. \ref{fig_fwnet}, we replace the fully-connected layers with convolutional layers, and operate $pool_p$ across all the output feature maps for each location individually. The latter is inspired by the well-known conversion between convolutional and fully-connected layers by reshaping inputs\footnote{\url{http://cs231n.github.io/convolutional-networks//\#convert}}, and turns $pool_p$ into a form of cross-channel parametric pooling \cite{goodfellow2013maxout,lin2013network,pfister2015flowing}.

The convolutional F-W Net then bears the similar flexibility to jointly learn weights and hyper-parameters ($p$ and $\gamma^t$). Yet different from the original version of F-W Net, the $pool_p$ in convolutional F-W Net reflects a diversified treatment of different convolutional filters at the same location, and should be differentiated from pooling over multiple points:
\begin{itemize}
	\item $p = 1$, only the channel containing the strongest response will be preserved at each location, which is reduced to max-out \cite{goodfellow2013maxout}.
	\item $p = 2$, $pool_p$ re-scales the feature maps across the channels by its root-mean-square.
	\item $p = \infty$ denotes the equal importance of all channels and leads to the cross-channel average.
\end{itemize}
By involving $p$ into optimization, $pool_p$ essentially learns to re-scale local responses, as advocated by the neural lateral inhibition mechanism. The learned $p$ will indicate the relative importance of different channels, and can potentially be useful for network compression \cite{han2015deep}.

The convolutional kernels in the convolutional F-W Net are not directly tied with any dictionary, thus we have not seen an initialization strategy straightforwardly available. In this paper, we use random initialization for the convolutional filters, but we initialize $p$ and $\gamma^t$ in the same way as mentioned before. As a result, convolutional F-W Nets often converge slower than fully-connected ones. We notice some recent works that construct convolutional filters \cite{chan2015pcanet,bruna2013invariant} from data in an explicit unsupervised manner, and plan to exploit them as future options to initialize convolutional F-W Nets.

To demonstrate its capability, we apply convolutional F-W Net to low-level image processing tasks, including image denoising and image super-resolution in this paper. Our focus lies on building light-weight compact models and verifying the benefit of introducing the learnable $p$.
\section{Experiments of Convolutional Frank-Wolfe Network}\label{sec_experiments}
\subsection{Image Denoising}
We investigate two versions of F-W Net for image denoising. The first version is the fully-connected (FC) F-W Net as used in the previous simulation. Here, the basic FC F-W Net (Fig. \ref{fig_fwnet}) is augmented with one extra fully-connected layer, whose parameters are denoted by $W_R\in \R^{n\times m}$ to reconstruct $\hat{\mathbf{x}}=W_R\mathbf{z}$. $W_R$ is naturally initialized by $D$. The network output is compared against the original/clean image to calculate MSE loss. To train this FC F-W Net, note that one strategy to image denoising is to split a noisy image into small (like $8\times 8$) overlapping patches, process the patches individually, and compose the patches back to a complete image. Then, we re-shape $8\times 8$ blocks into 64-dim samples (i.e. $n = 64$). We adopt $m=512$, $T=8$, and $c=1$. The network is trained with a number of $8\times 8$ noisy blocks as well as their noise-free counterparts.
The second version of F-W Net is the proposed convolutional (Conv) F-W Net as discussed in Section \ref{sec_convfwnet}. The adopted configurations are: $3\times 3$ filters, 64 feature maps per layer, $T=4$, and $c=1$.

We use the BSD-500 dataset \cite{arbelaez2011contour} for experiment. The BSD-68 dataset is used as testing data, and the remaining 432 images are converted to grayscale and added with white Gaussian noise $\mathcal{N}(0, \sigma^2)$ for training.
Several competitive baselines are chosen: a FC LISTA \cite{gregor2010learning} network that is configured identically to our FC F-W Net; a Conv LISTA network that is configured identically to our Conv F-W Net; KSVD + OMP that represents the $L_0$-norm optimization; BM3D \cite{burger2012image} using the dictionary size of 512; and DnCNN \cite{zhang2017beyond} which is a recently developed method based on CNN, here our re-trained DnCNN includes 4 convolutional layers followed by BN and ReLU. We consider three noise levels: $\sigma=15$, $\sigma=25$, and $\sigma=50$.

Table \ref{tab_den} provides the results of different image denoising methods on the BSD-68 dataset. FC F-W Net is better than LISTA in all cases. We also study the effect of $p$, as shown in Table \ref{tab_den_p}. F-W Net with learnable $p$ outperforms F-W Net with fixed $p=2$ by a large margin, and is slightly better than F-W Net with fixed $p=1.4$.
Here, learnable $p$ seems benefiting not only the final model but also the training itself, as has been observed in \ref{sec_mnist}.
BM3D is a strong baseline, which the deep networks cannot beat easily. As seen from Table \ref{tab_den}, BM3D outperforms DnCNN (4 layers), but Conv F-W Net (4 layers) is better than BM3D.

\begin{figure}
\centering
\includegraphics[width=.35\textwidth]{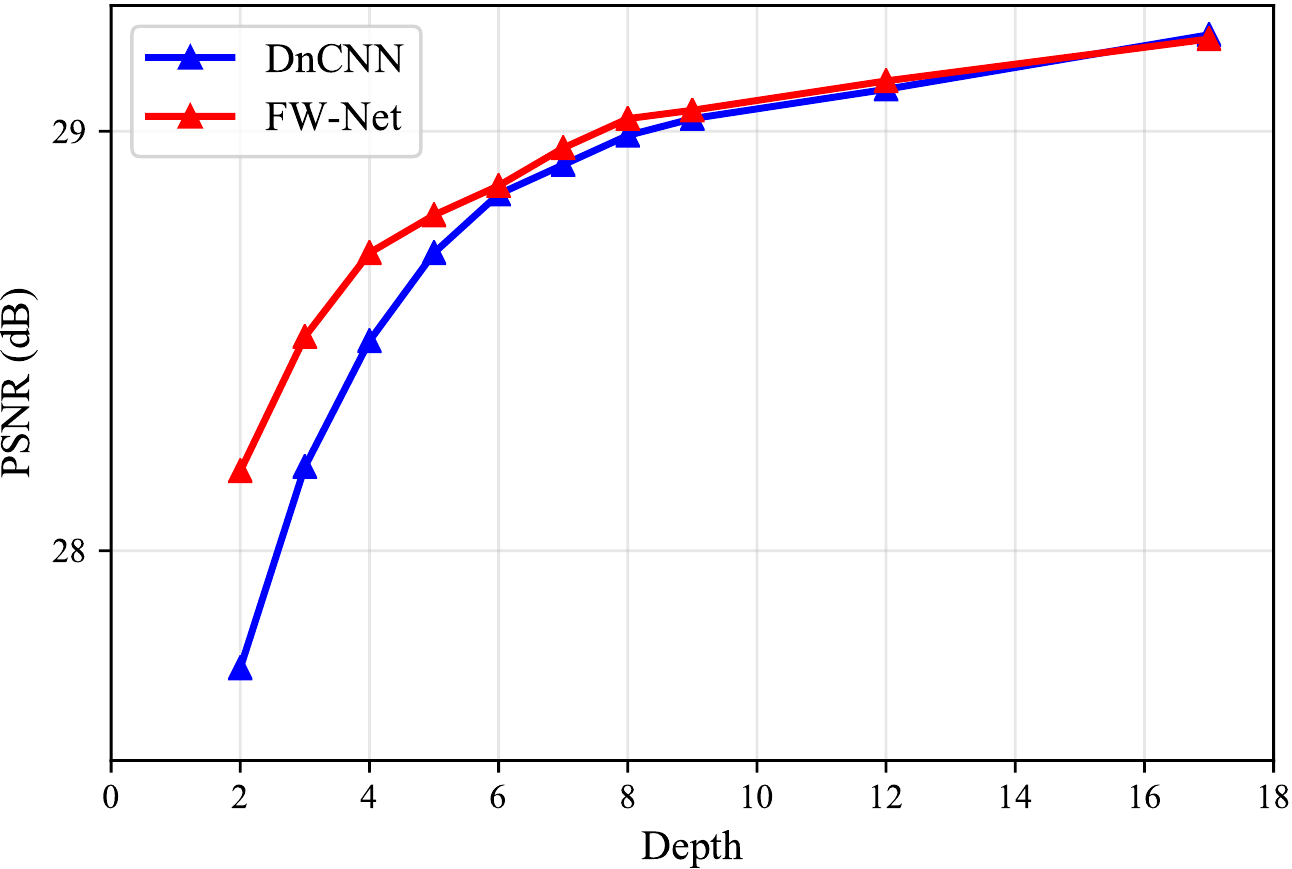}
\includegraphics[width=.35\textwidth]{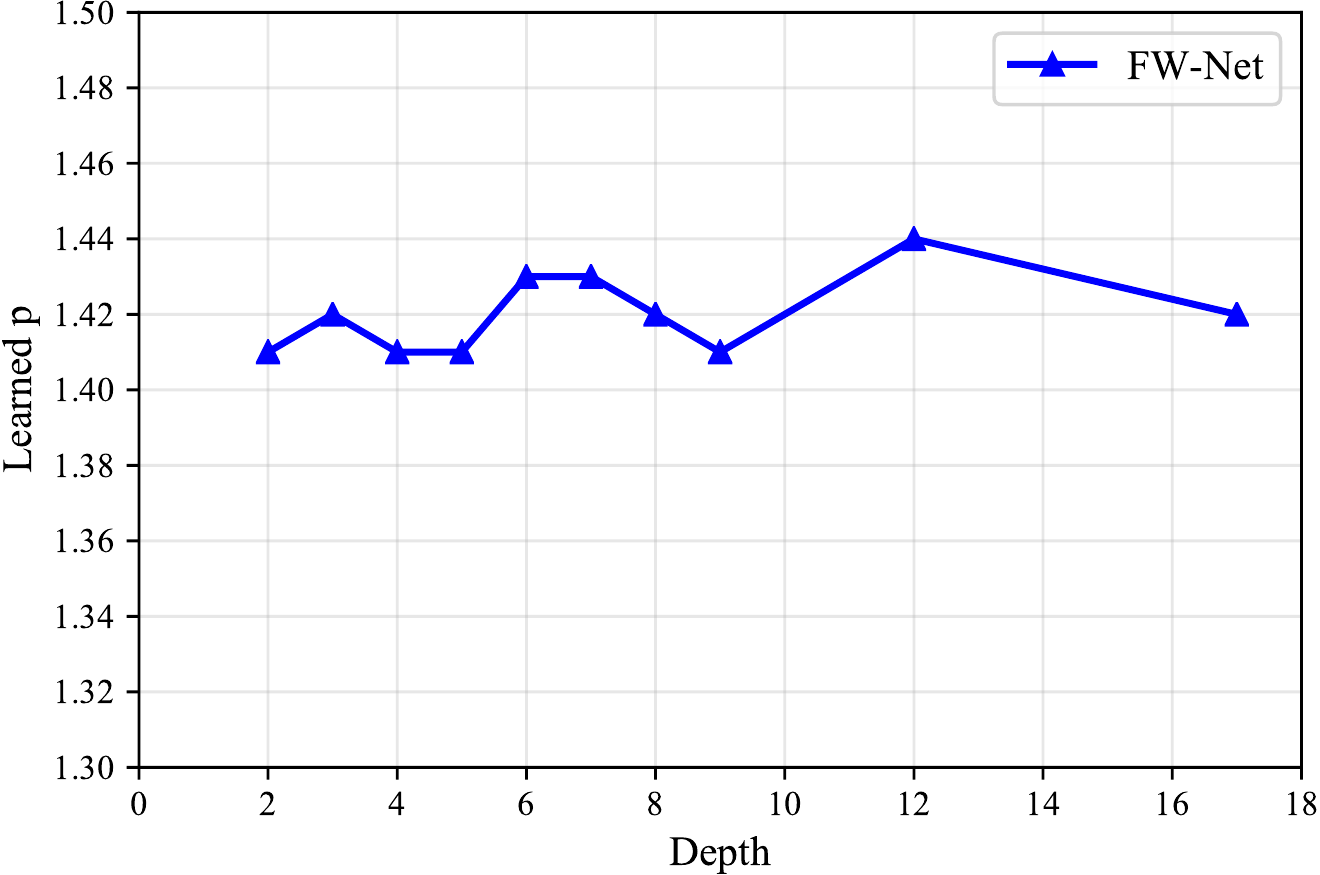}
\caption{Top: image denoising results on the (gray) BSD-68 dataset when $\sigma=25$, with DnCNN and (Conv) F-W Net at different depths. Bottom: learned $p$ value of the Conv F-W Net at different depths.}
\label{denoise_diff_depth}
\end{figure}

It is worth noting that the original DnCNN in \cite{zhang2017beyond} has 20 layers and much more parameters and outperforms BM3D. We conduct an experiment to compare DnCNN with Conv F-W Net at different number of layers. The results are shown in Fig. \ref{denoise_diff_depth} (a). We observe that for shallow networks, Conv F-W Net outperforms DnCNN significantly. As the network depth increases, both Conv F-W Net and DnCNN perform similarly. This may be attributed to the learnable parameter $p$ in the Conv F-W Net, which helps much when the network parameters are less. Thus, $pool_p$ may be favorable if we want to build a compact network. The learned $p$ values are shown in Fig. \ref{denoise_diff_depth} (b). We observe that the learned $p$ value is stable across networks with different depths. It is also similar to the learned $p$ value in the FC F-W Net ($p=1.41$). Thus, we consider that the $p$ value is determined by the data, and F-W Net can effectively identify the $p$ value regardless of FC or Conv structures.
\begin{table}
\centering
\caption{Image denoising results (average PSNR/dB) on the (gray) BSD-68 dataset. All the networks have 4 fully-connected or convolutional layers for fair comparison.}\label{tab_den}

\begin{tabular}{l|ccc}
	\hline
	& $\sigma=15$& $\sigma=25$&$\sigma=50$\\
	\hline
	FC LISTA & 29.20 & 27.63 & 24.33  \\
	FC F-W Net, learned $p=1.41$ & 29.35 & 27.71 & 24.52 \\
	\hline
  KSVD + OMP & 30.82 & 27.97 & 23.97 \\
	BM3D 	  &  31.07 & 28.57 & 25.62 \\
  \hline
	DnCNN (4 layers) & 30.89 & 28.42 & 25.42 \\
  DnCNN (4 layers, w/o BN) & 30.85 & 28.43 & 25.36 \\
  Conv LISTA (4 layers) & 30.90 & 28.50 & 25.55 \\
	Conv F-W Net (4 layers) & \textbf{31.27} & \textbf{28.71} & \textbf{25.66} \\
	\hline
\end{tabular}
\end{table}
\begin{table}
\centering
  \caption{Results of FC F-W Nets with fixed $p$ and learnable $p$ for image denoising on the BSD-68 dataset.}\label{tab_den_p}

\begin{tabular}{l|ccc}
	\hline
	& $\sigma=15$& $\sigma=25$&$\sigma=50$\\
	\hline
	F-W Net, fixed $p=2$ & 28.35 & 26.41 & 23.68 \\
	F-W Net, fixed $p=1.4$ & 29.25 & 27.62 & 24.45 \\
	F-W Net, learned $p=1.41$ & \textbf{29.35} & \textbf{27.71} & \textbf{24.52} \\
	\hline
\end{tabular}
\end{table}
\subsection{Image Super-Resolution}
For the experiments about single image super-resolution (SR), we compare Conv F-W Net with baselines SRCNN \cite{dong2016image} and VDSR \cite{kim2016accurate}, and all methods are trained on the 91-image standard set and evaluated on Set-5/Set-14 with a scaling factor of 3. We train a 4-layer Conv F-W Net and a 4-layer VDSR (4 convolutional layers equipped with ReLU), both of which have the same number of convolutional kernels. We adopt the same training configurations as presented in \cite{kim2016accurate}. For SRCNN we directly use the model released by the authors.

Table \ref{tab_sr} compares the results of these methods. Our Conv F-W Net achieves the best performance among the three. To further understand the influence of each component in F-W Net, we experimentally compare the following setttings:
\begin{itemize}
	\item F-W Net (No Skip): removing the top skip connections in Fig. \ref{fig_fwnet};
	\item F-W Net (Fixed $\gamma=1$): setting $\gamma^t=1,t= 1,2, \dots T-1$, which is equivalent to removing the bottom skip connections in Fig. \ref{fig_fwnet};
	\item F-W Net (ReLU): replacing all the $pool_p$ units with ReLU.
\end{itemize}
Both \textit{F-W Net (No Skip)} and \textit{F-W Net (Fixed $\gamma=1$)} break the original structure introduced by the Frank-Wolfe algorithm, and incur severe performance drop. \textit{F-W Net (ReLU)} performs similarly to 4-layer VDSR and is worse than F-W Net. These results further demonstrate that each component in Conv F-W Net contributes to the final performance.

We also measure the effect of the hyper-parameter $p$. The results are shown in Table \ref{tab_sr_ablation}. Different $p$ values indeed influence on the final performance significantly, and F-W Net with learnable $p$ achieves the best performance, which again verifies the advantage of attaining the prior from the data.
\begin{table}
\centering
\caption{$3\times$ image super-resolution results (average PSNR/dB) on the (gray) Set-5 and Set-14 datasets.}\label{tab_sr}

	\begin{tabular}{l|cc}
	\hline
	Method& Set-5 & Set-14 \\
	\hline
	3-layer SRCNN & 32.34 & 28.64 \\
	4-layer VDSR & 32.51 & 28.71 \\
  4-layer VDSR (+BN) & 32.55 & 28.69 \\
	
	4-layer Conv F-W Net  & \textbf{32.85} & \textbf{28.76} \\
	\hline
	\end{tabular}
\end{table}
\begin{table}
\centering
  \caption{Ablation study results of Conv F-W Net for image super-resolution on the Set-5 and Set-14 datasets.}\label{tab_sr_ablation}

	\begin{tabular}{l|cc}
	\hline
	Method& Set-5 & Set-14 \\
	\hline
	F-W Net (No Skip)  & 31.52 & - \\
	F-W Net (Fixed $\gamma=1$)  & 32.21 & - \\
	F-W Net (ReLU)  & 32.57 & 28.66 \\
	F-W Net         & \textbf{32.85} & \textbf{28.76} \\
	\hline
	F-W Net (Fixed $p=1.3$)  & 32.49 & 28.42 \\
	F-W Net (Fixed $p=1.5$)  & 32.81 & 28.63 \\
	F-W Net (Fixed $p=1.8$)  & 32.73 & 28.69 \\
	F-W Net (Fixed $p=2.3$)  & 32.59 & 28.58 \\
	F-W Net (Fixed $p=2.5$)  & 32.65 & 28.67 \\
	F-W Net (Learned $p=1.489$)  & \textbf{32.85} & \textbf{28.76} \\
	\hline
	\end{tabular}
\end{table}
\section{Conclusion}\label{sec_conclusion}
We have studied the general non-sparse coding problem, i.e. the $L_p$-norm constrained coding problem with general $p>1$.
We have proposed the Frank-Wolfe network, whose architecture is carefully designed by referring to the Frank-Wolfe algorithm. Many aspects of F-W Net are inherently connected to the existing success of deep learning. F-W Net has gained impressive effectiveness, flexibility, and robustness in our conducted simulation and experiments. Results show that learning the hyper-parameter $p$ is beneficial especially in real-data experiments, which highlights the necessity of introducing general $L_p$-norm and the advantage of F-W Net in learning the $p$ during the end-to-end training.

Since the original Frank-Wolfe algorithm deals with convex optimization only, the proposed F-W Net can handle $p\geq 1$ cases, but not $p<1$ cases. Thus, F-W Net is good at solving non-sparse coding problems. $p=1$ is quite special, as it usually leads to sparse solution \cite{donoho2006most}, thus, F-W Net with fixed $p=1$ can solve sparse coding, too, but then its efficiency seems inferior to LISTA as observed in our experiments. For a real-world problem, is sparse coding or non-sparse coding better? This is an open problem and calls for future research. In addition,
a number of promising directions have emerged as our future work, including handling more constraints other than the $L_p$-norm, and the customization of F-W Net for more real-world applications.

\ifCLASSOPTIONcaptionsoff
  \newpage
\fi



\end{document}